\documentclass[final,5p,times]{elsarticle}
\usepackage{graphicx}
\usepackage{lineno}

\usepackage{subfigure}
\usepackage{epsfig}     % for postscript graphics files
\usepackage{amsmath}    % assumes amsmath package installed
\usepackage{amsthm}
\usepackage{amssymb}    % assumes amsmath package installed
\usepackage{algorithmic}
\usepackage{makecell,multirow,diagbox}
\usepackage{booktabs}
\usepackage{hyperref}
\usepackage{caption}
\hypersetup{
	colorlinks=true,
	linkcolor=blue,
	citecolor=green}
\graphicspath{{figures/}}

\journal{Neural Networks}

\begin{document}
\begin{sloppypar}
\begin{frontmatter}




\title{Recurrent Neural Network from Adder's Perspective: \\Carry-lookahead RNN}


%\author{}

%\address{}
\author{Haowei Jiang\texorpdfstring{$^a$}{a}, Feiwei Qin\texorpdfstring{$^a$}{a}, Jin Cao\texorpdfstring{$^{b,*}$}{b,*}, Yong Peng\texorpdfstring{$^a$}{a}, Yanli Shao\texorpdfstring{$^a$}{a}\\
\texorpdfstring{$^a$}{a}School of Computer Science and Technology, Hangzhou Dianzi University, China\\
\texorpdfstring{$^b$}{b}Whiting School of Engineering, Johns Hopkins University, USA}

\begin{abstract}
The recurrent network architecture is a widely used model in sequence modeling, but its serial dependency hinders the computation parallelization, which makes the operation inefficient. The same problem was encountered in serial adder at the early stage of digital electronics. In this paper, we discuss the similarities between recurrent neural network (RNN) and serial adder. Inspired by carry-lookahead adder, we introduce carry-lookahead module to RNN, which makes it possible for RNN to run in parallel. Then, we design the method of parallel RNN computation, and finally Carry-lookahead RNN (CL-RNN) is proposed. CL-RNN takes advantages in parallelism and flexible receptive field. Through a comprehensive set of tests, we verify that CL-RNN can perform better than existing typical RNNs in sequence modeling tasks which are specially designed for RNNs. Code and models are available at: \url{https://github.com/WinnieJiangHW/Carry-lookahead_RNN}.
\end{abstract}


\end{frontmatter}



\section{Introduction}\label{sec-introduction}

Recurrent network architecture is a common model used to handle sequence tasks. The title of sequence modeling chapter in the canonical textbook \emph{Deep Learning}~\cite{goodfellow2016deep} is "Sequence Modeling: Recurrent and Recursive Nets". Recurrent neural network (RNN) proposed by Rumelhart et al.~\cite{rumelhart1986learning} and its improved models such as long short-term memory (LSTM)~\cite{hochreiter1997long}, gated recurrent neural networks~\cite{chung2014empirical}, are firmly established to deal with language modeling, machine translation, etc.~\cite{sutskever2014sequence, bahdanau2014neural, cho2014learning} Numerous researches on recurrent network architecture push its boundaries and application in natural language processing~\cite{wu2016google, jozefowicz2016exploring}.

However, recurrent neural networks rely on the preorder state $h_{t-1}$ to generate the output and postorder state $h_{t}$. This serial dependency hinders the parallelization of computation, which makes the model run inefficient.

In the history of digital electronics, the adder as the most basic computing unit has faced the same problem. The traditional serial adder has to wait for the completion of the preorder computation in order to obtain the carry bit. Then it generates next carry bit and delivers. Considering the serial adder is analogous to the recurrent network architecture in various aspects, we suppose that there is an analogy between them.

Digital electronics has developed into a mature subject nowadays. Meanwhile, parallelization improvement has always been a research hotspot in the discipline~\cite{kirichenko2019ersfq, thapliyal2005novel, ZHOU2018473, hou2019survey, zhou2018accelerating}. As one of the most fundamental components, adder has been deeply studied. Given the analogy between adder and recurrent network architecture, we suppose that it is valid to apply the proven optimization experience on adder to improving recurrent neural network.

By extension, we suppose that the knowledge of digital electronics could be informative to the design of neural network architectures. Neural network architectures have commonalities with electronic circuits in terms of structure and other characteristics. Based on that, the proven experience on digital electronics can be applied to improving neural networks. For example, shortcut connection used in ResNet~\cite{he2016deep} is similar to short circuit, which is a common method used in electronic circuits; branch structure adopted by GoogLeNet~\cite{szegedy2015going} is also a basic structure in electronic circuits.

Carry-lookahead adder~\cite{rosenberger1960simultaneous} is today's mainstream adder. It innovatively proposes the carry-lookahead module and applies independent formulas to enable the parallel calculation of multi-bit. Compared with the traditional serial adder, it greatly improves the computation efficiency.

Inspired by carry-lookahead adder, we introduce carry-lookahead module to RNN. Furthermore, the method of parallel RNN computation is designed and finally Carry-lookahead RNN (CL-RNN) is proposed.

Through a comprehensive set of tests, CL-RNN not only achieves parallelization compared to recurrent neural networks, but also shows stronger representation capability in sequence modeling.

In summary, the main contributions of this paper are as follows.

\begin{enumerate}\setlength{\itemsep}{4pt}
	\item[1)] Consider the similarities between RNN and serial adder in various aspects. Taking their analogy as the core point, we propose the idea of transferring the optimization experience on adder to RNN.
	\item[2)] Inspired by carry-lookahead adder, we propose Carry-lookahead RNN, which consists of carry-lookahead module and parallel RNN module. The carry-lookahead module provides support for model’s parallelization, and the parallel RNN module improves the representation capability and eases the model optimization difficulty.
	\item[3)] Compare CL-RNN with classic RNNs as well as TCN. Through ablation experiment, we verify the effectiveness of the model improvement. After a series of stress tests, it's found that CL-RNN can perform better than classic TCN, RNN and LSTM in sequence modeling, which recurrent network architecture excel in.
\end{enumerate}

\section{Related Work}\label{sec-related_work}

Since the classic recurrent neural network has been proposed by Rumelhart in 1986~\cite{rumelhart1986learning}, a lot of related researches are conducted and many variants are proposed. Recurrent neural networks now have been a large family, and extensively applied in language modeling, machine translation, etc.~\cite{sutskever2014sequence, bahdanau2014neural, cho2014learning}

Some of the classic variants greatly improves the performance of RNN. Long short-term memory~\cite{hochreiter1997long} introduces the structure of "gates" to RNN, which allows RNN to forget irrelevant information and keep important information in long-term; Gate Recurrent Unit~\cite{chung2014empirical} simplify LSTM to some extent, which accelerate the training efficiency.

In recent years, more and more improved models are proposed. krueger et al. regularize the penalty term of LSTM and propose Norm-stabilized LSTMs~\cite{krueger2015regularizing}; Vohra et al. superimpose Deep Belief Net~\cite{hinton2009deep} on LSTM and propose DBN LSTM~\cite{vohra2015modeling}, which can perform very well on certain datasets. Other recent works combine RNN with CNN, such as Convolutional LSTM~\cite{shi2015convolutional}, which replaces the fully-connected layer in LSTM with CNN; Quasi-RNN~\cite{bradbury2016quasi}, which interleaves convolutional layers with recurrent layers; Dilated RNN~\cite{chang2017dilated}, which introduces dilation into recurrent network architecture.

The various variants proposed make the performance of RNN improve a lot. However, RNNs are still limited by its serial dependency, which seriously lowers the computation efficiency and hinders its large-scale applications.

Although some recent works have somewhat optimized the efficiency through factorization techniques~\cite{kuchaiev2017factorization}, conditional operations~\cite{shazeer2017outrageously}, etc., the fundamental constraints of sequential computation still exist and limit its further development.

On the other hand, some researches try to use convolutional neural network (CNN)~\cite{lecun1989backpropagation, krizhevsky2012imagenet} to deal with sequence tasks, because CNN is easy to run in parallel. Temporal convolutional network (TCN), which mainly refers to the Dilated TCN proposed in~\cite{lea2017temporal}, is a paradigm of using CNN for sequence modeling. Compared with classic CNN, temporal convolutional network has two distinctive features:

\begin{enumerate}
	\item[1)] By causal convolution~\cite{oord2016wavenet}, each neuron can only receive the input sequence before its corresponding word, which means there is no possibility of information "leakage" from the future to the past.
	\item[2)] It can receive input sequence of any length and map them to output sequence of the same length, which is consistent with RNN.
\end{enumerate}

While classic TCN is proposed to process video data, new researches have found that TCN can also provide effective support for language modeling, with results comparable to mature sequence models such as LSTM, Gate Recurrent Unit~\cite{chung2014empirical, bai2018empirical, oord2016wavenet, kalchbrenner2016neural}. Among the works, Bai et al. proposes a variant of TCN~\cite{bai2018empirical}, which introduces residual structure to TCN, and achieves great performance in sequence modeling tasks.

\section{Model}\label{sec-model}

\begin{figure*}[htbp]
	\centering
	\includegraphics[width=0.90\textwidth]{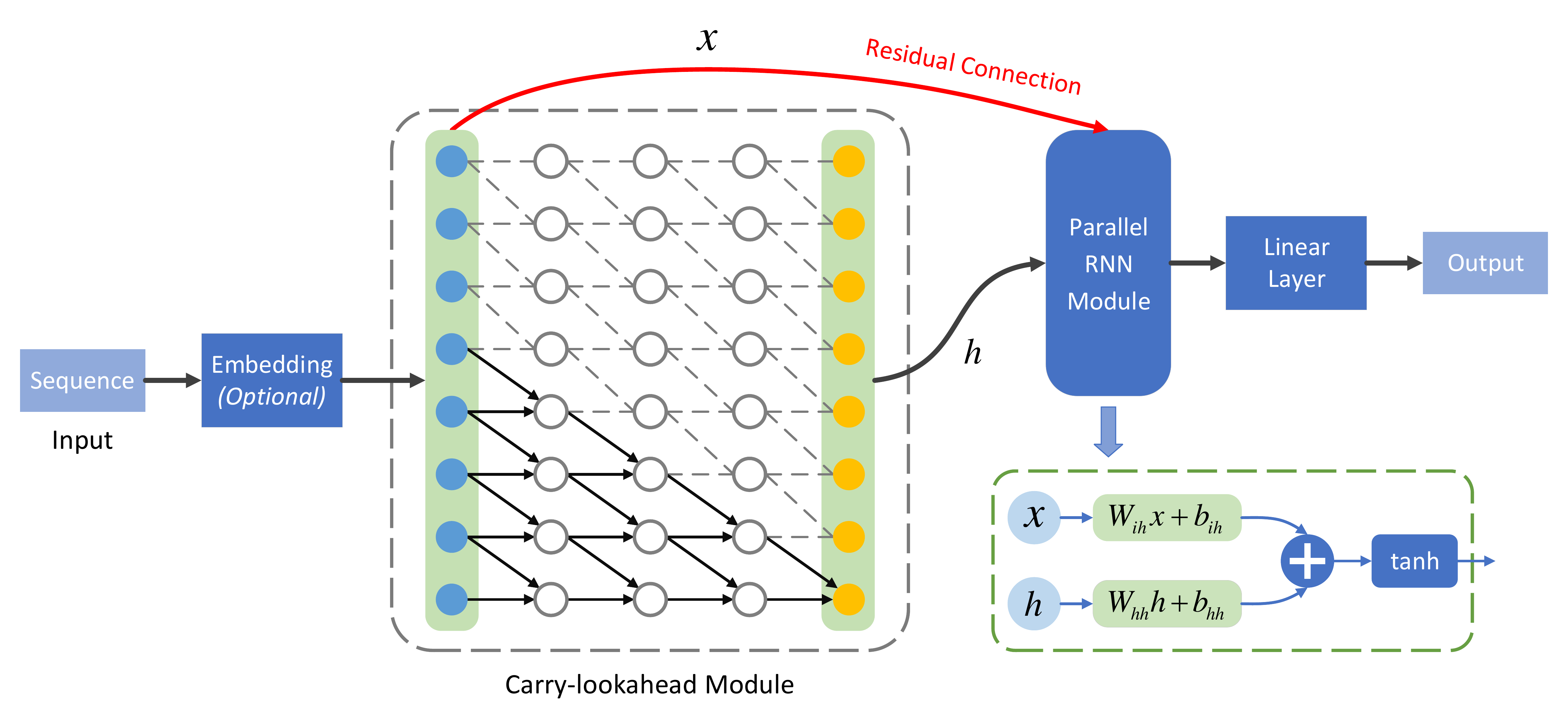}
	\caption{Overview of Carry-lookahead RNN. The model receives sequence data as input. Embedding layer is optional depending on the type of input sequence. Carry-lookahead module generates hidden states $h$ in advance. Parallel RNN module receives both original input $x$ (residual connection marked in red) and hidden states $h$. A linear layer is added in the end to generate the output.}\label{Model-structure}
\end{figure*}

In this section, we first discuss the similarities between serial adder and RNN. Then, inspired by carry-lookahead adder, the carry-lookahead module is introduced to RNN, which makes it possible for RNN to run in parallel. Also, we design the method of parallel RNN computation, and finally Carry-lookahead RNN (CL-RNN) is proposed.

The structure of our proposed model is shown in Fig.~\ref{Model-structure}. The core part of CL-RNN consists of two modules: Carry-lookahead Module and Parallel RNN Module. Sequence input is firstly sent to carry-lookahead module to calculate hidden states in advance. Then parallel RNN module receives both hidden states and original sequecne input to generate the output. Meanwhile, parallel RNN module plays a role as residual component for the overall model, which eases the model optimization difficulty.

\subsection{Motivation}

\begin{figure}[htbp]
	\centering
	\subfigure[\label{serial-adder}]{\includegraphics[width = 0.45\textwidth]{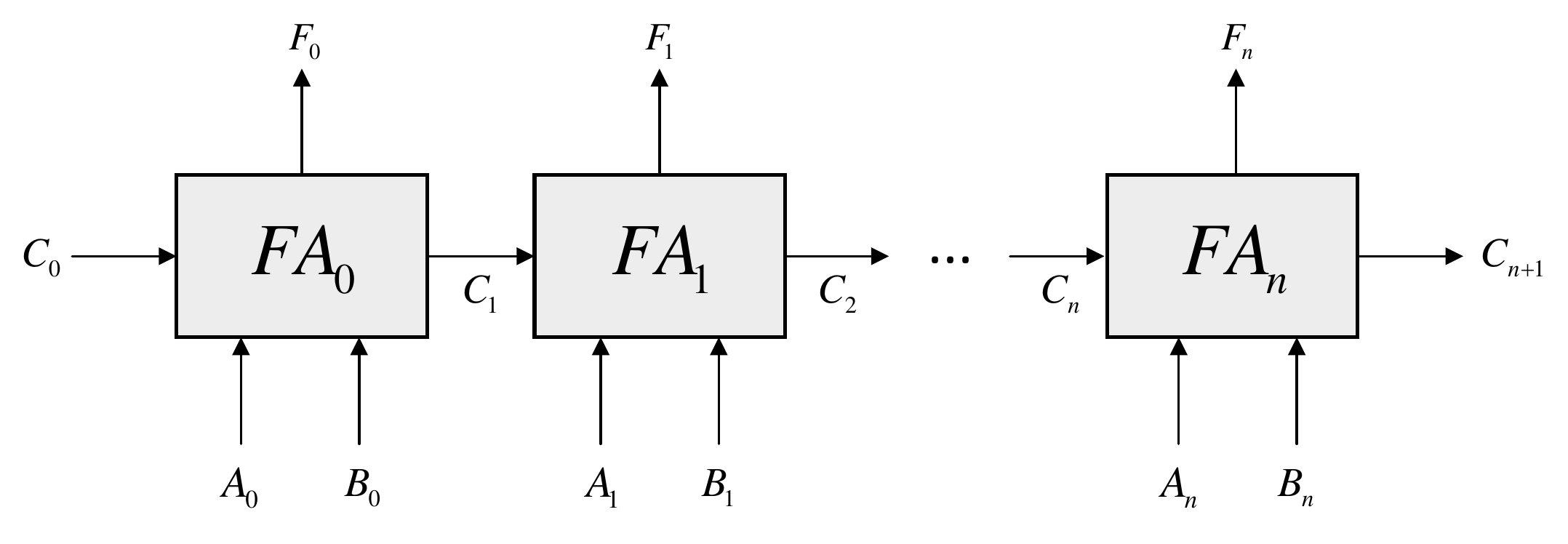}}
	\\
	\centering
	\subfigure[\label{RNN}]{\includegraphics[width = 0.40\textwidth]{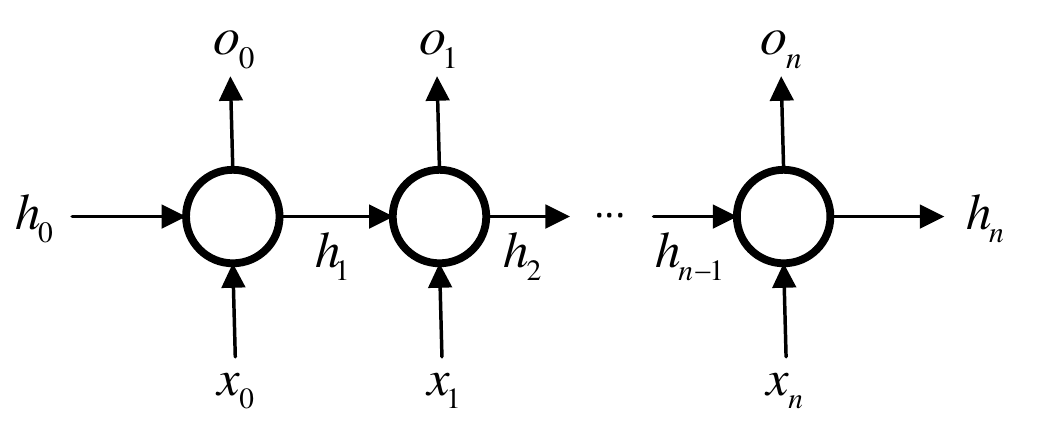}}
	\caption{The structure of (a) serial adder and (b) RNN. They are typical recurrent structure in their respective fields and similar in various aspects.}
\end{figure}

Serial adder consists of several full adders connected in series, and the structure is shown in Fig.~\ref{serial-adder}. Each full adder $FA_i$ waits and receives the carry bit $C_i$ generated by the previous full adder $FA_{i-1}$, and generates the next carry bit $C_{i+1}$ recursively.

Recurrent neural network unfolds with a similar structure to the serial adder, as shown in Fig.~\ref{RNN}. Each neuron receives the hidden state $h_{t-1}$ generated by the previous neuron. Then it produces the result $o_t$ and the new hidden state $h_t$. Noting that the internal parameters of each neuron in RNN are shared, which means the internal function is same. This is consistent with the fact that each full adder in serial adder has the same operation function.

Their similar structures lead to the same problem, i.e., serial dependency hinders parallel computation and lowers efficiency in large-scale applications. Due to their similarities in various aspects, we consider them analogous. Since the optimization for adder is a mature topic, the optimization experience on adder can be applied to improving recurrent neural network.

\subsection{Carry-lookahead Module}\label{sec-clm}

\begin{figure*}[htbp]
	\centering
	\subfigure[\label{carry-lookahead-RNN}]{\includegraphics[width = 0.435\textwidth]{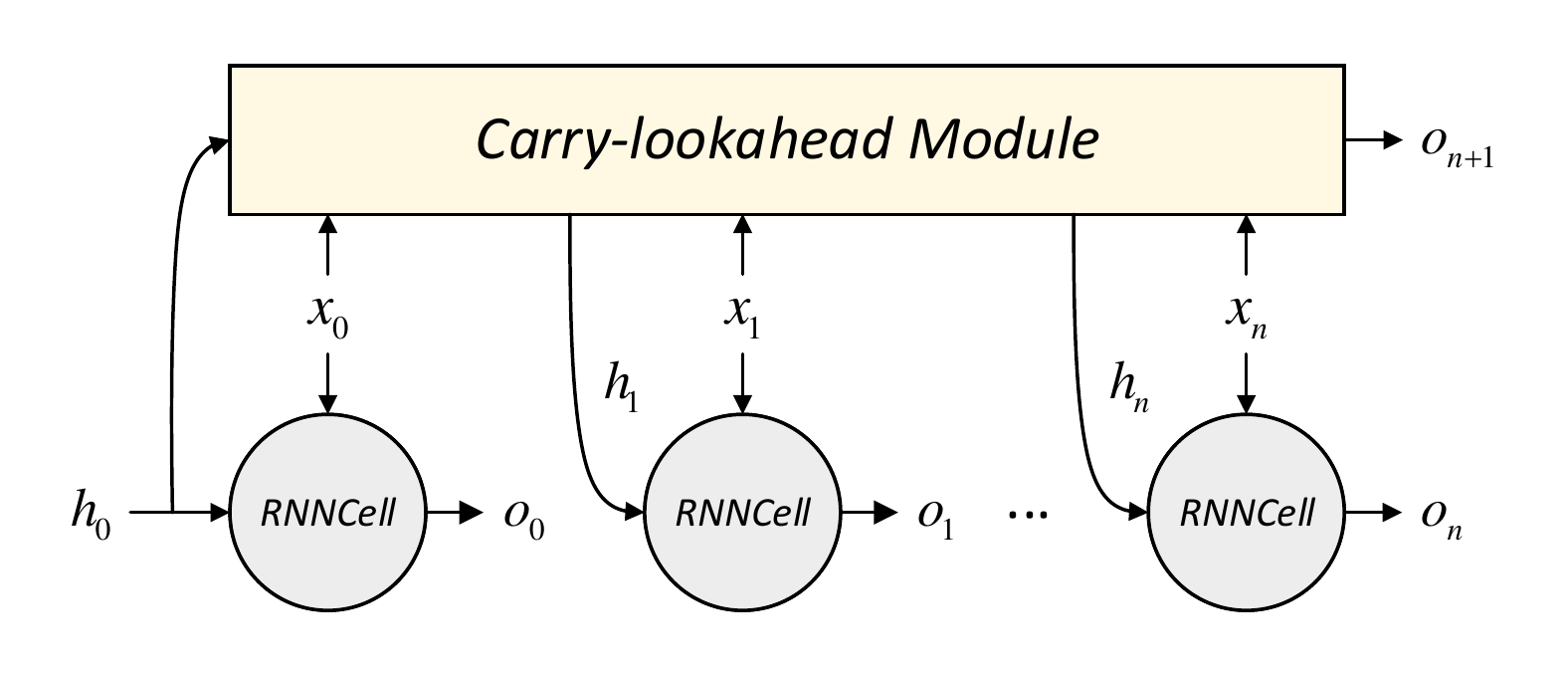}}
	\subfigure[\label{carry-lookahead-adder}]{\includegraphics[width = 0.465\textwidth]{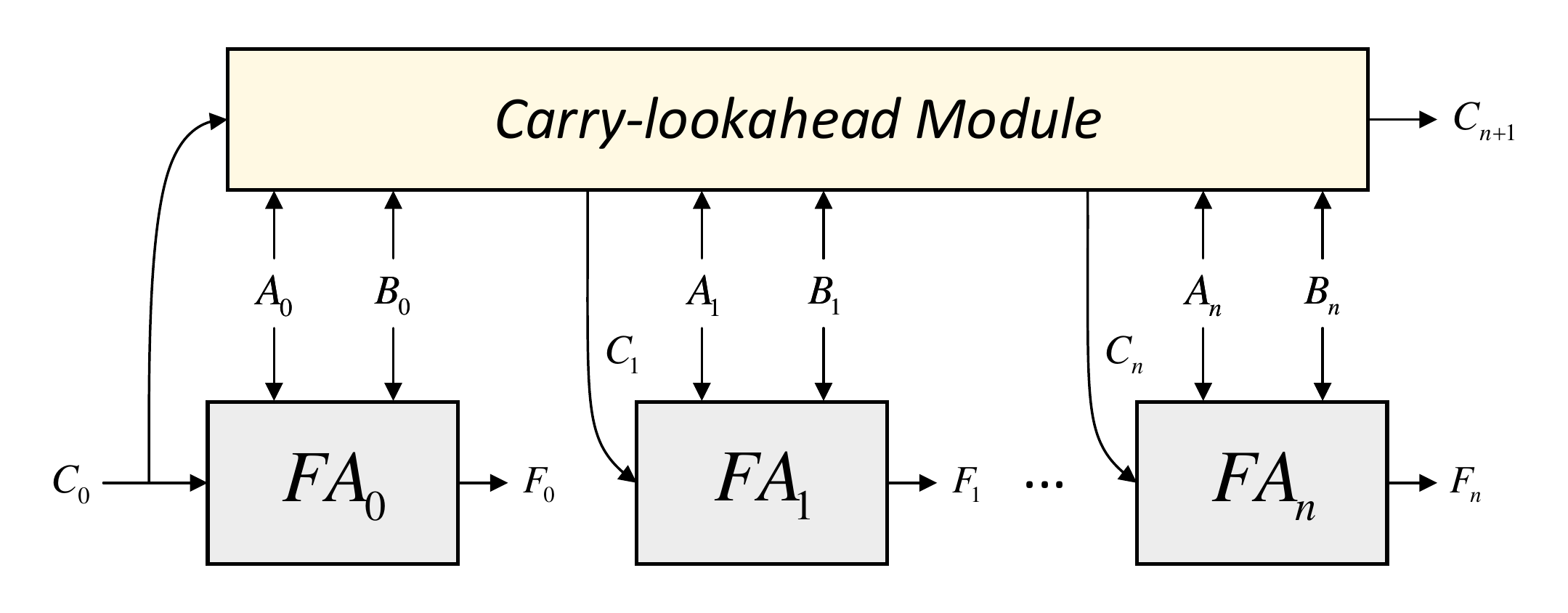}}
	\\
	\centering
	\subfigure[\label{carry-lookahead-module}]{\includegraphics[width = 0.6\textwidth]{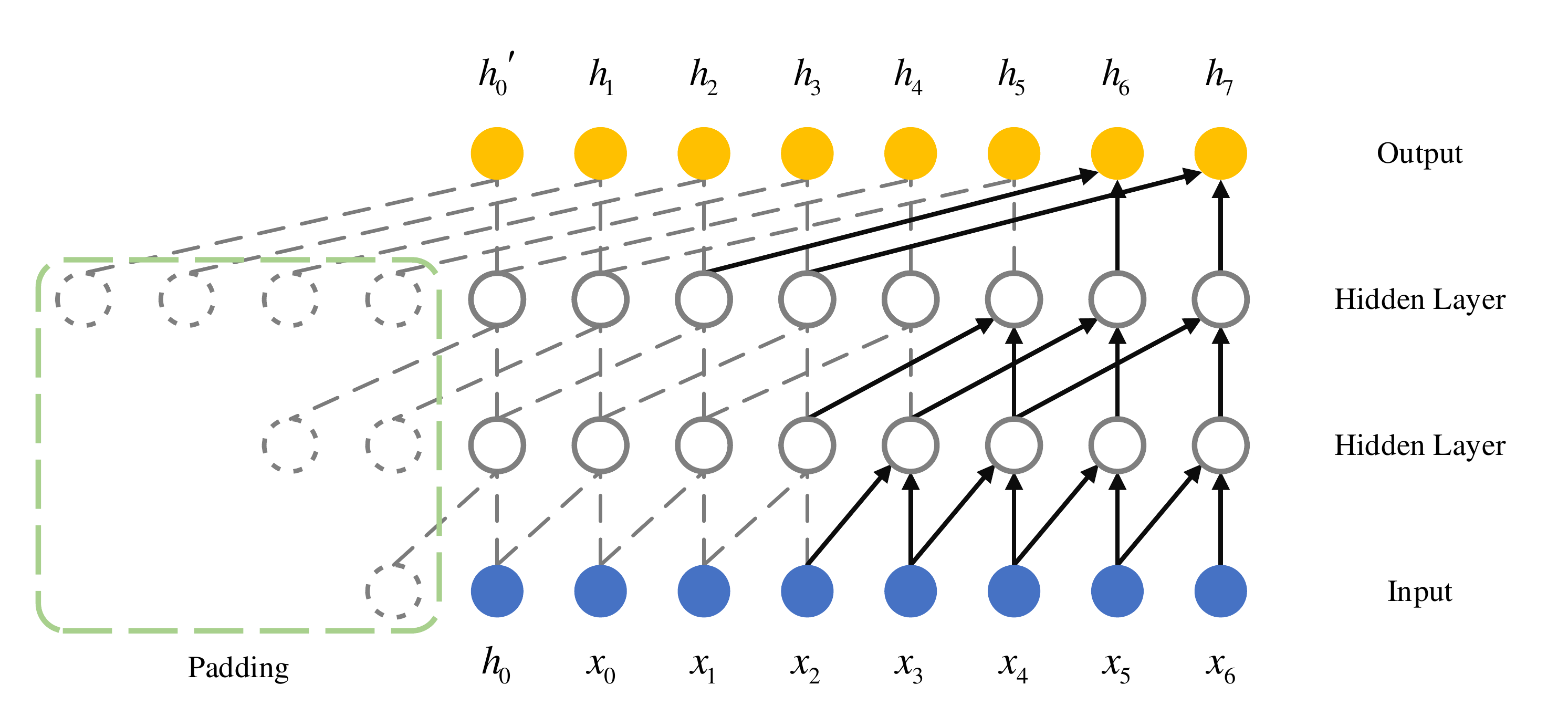}}
	\subfigure[\label{carry-lookahead-block}]{\includegraphics[width = 0.3\textwidth]{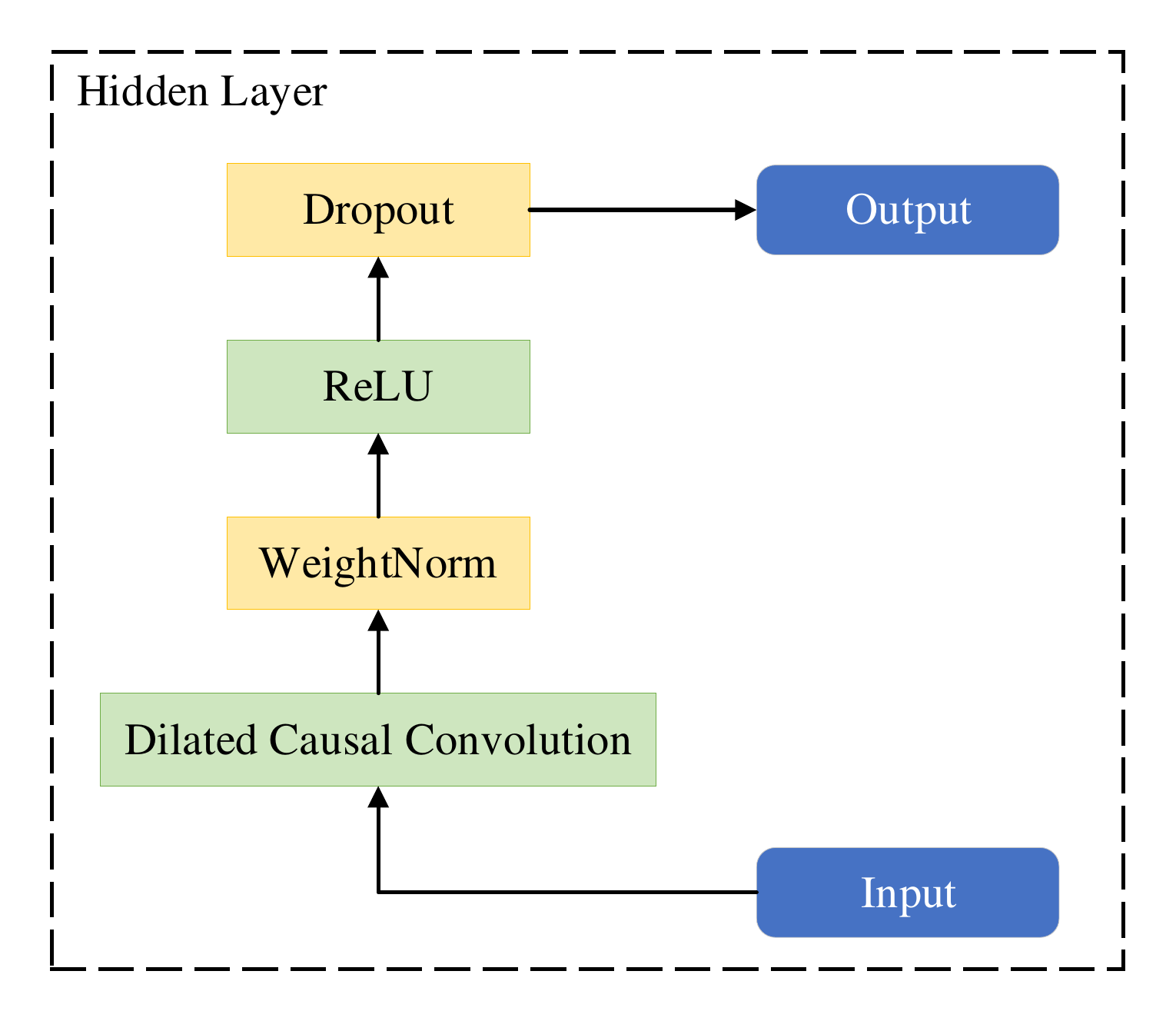}}
	\caption{(a) An adder-style edition of Carry-lookahead RNN, which hides the internal implementation. (b) The structure of carry-lookahead adder. (c) The structure of Carry-lookahead Module with dilation factor $d$ = 1, 2, 4. (d) The implementation of the hidden layer in (c).}
\end{figure*}

Carry-lookahead adder is an improved version of serial adder, which is the mainstream adder nowadays. It consists of a carry-lookahead module and a number of full adders. The structure is shown in Fig.~\ref{carry-lookahead-adder}, and the basic principle of carry-lookahead module is shown in Equation (\ref{equ-carry-lookahead-module}).

\begin{equation}\label{equ-carry-lookahead-module}
	\begin{cases}
		G_i = {A_i \cdot B_i}\\
		P_i = {A_i + B_i}\\
		C_{i+1} = {G_i + P_i \cdot C_i}\\
	\end{cases}
\end{equation}

Taking 4-bit adder as an example, Equation (\ref{equ-carry-lookahead-module-1}) can be derived based on Equation (\ref{equ-carry-lookahead-module}), so that the carry bit of each full adder can be calculated in advance. Carry-lookahead module breaks the serial dependency between full adders, realizes the parallelization of multi-bit addition, and greatly improves the efficiency.

\begin{equation}\label{equ-carry-lookahead-module-1}
	\resizebox{.91\hsize}{!}
	{$
		\begin{cases}
			C_1 = {G_0 + P_0 \cdot C_0}\\
			C_2 = {G_1 + P_1 \cdot G_0 + P_1 \cdot P_0 \cdot C_0}\\
			C_3 = {G_2 + P_2 \cdot G_1 + P_2 \cdot P_1 \cdot G_0 + P_2 \cdot P_1 \cdot P_0 \cdot C_0}\\
			C_4 = {G_3 + P_3 \cdot G_2 + P_3 \cdot P_2 \cdot G_1 + P_3 \cdot P_2 \cdot P_1 \cdot G_0 + P_3 \cdot P_2 \cdot P_1 \cdot P_0 \cdot C_0}\\
		\end{cases}
	$}
\end{equation}

Through Equation (\ref{equ-carry-lookahead-module-1}), the core idea of carry-lookahead module can be concluded as to calculate each carry bit with respective preorder input, which can be expressed as:

\begin{equation}\label{equ-carry-lookahead-module-2}
	C_{i+1} = {f_i(C_0, A_0, A_1, \cdots, A_i, B_0, B_1, \cdots, B_i)}
\end{equation}

Based on the principle, we introduce carry-lookahead module to RNN (Fig.~\ref{carry-lookahead-RNN}). Carry-lookahead module can be expressed as:
\begin{equation}
	h_{i+1} = 1D\underline{~~}CNN_i(h_0, x_0, \cdots, x_i)
\end{equation}
which is consistent with Equation (\ref{equ-carry-lookahead-module-2}) in form. It's implemented with multiple \textbf{dilated causal convolutional} blocks, and the structure is shown in Fig.~\ref{carry-lookahead-module}.

Dilated causal convolution combines the techniques of causal convolution and dilated convolution. They have their own unique purpose. \textbf{Causal convolution} avoids the information "leakage" from the future to the past, which enables carry-lookahead module to process sequential data effectively. \textbf{Dilated convolution} exponentially enlarges the receptive field. As a result, a very large receptive field
can be achieved with a few layers, which prevents the model from over-fitting the training data. The receptive field on each layer can be determined as:
\begin{equation}
	Receptive\underline{~~}Field(l) = \frac{k-1}{m-1}\times(m^l-1)+1
\end{equation}
where $l$ is level of layer, $k$ is kernel size, $m$ is multiple of dilation factor. According to the formula, receptive field can be flexibly adjusted by modifying dilation factor or the number of layers.

In a word, carry-lookahead module is the key component to achieve parallelism. It generates hidden states in advance to replace the “recurrent” computation in RNN. Besides, it has the good characteristics that there is no maximum length limitation of input, and the output sequence of the module is the same length as the input sequence, which provides convenience for subsequent operation.

\subsection{Parallel RNN Module}

\begin{figure*}[htbp]
	\centering
	\includegraphics[width=1.00\textwidth]{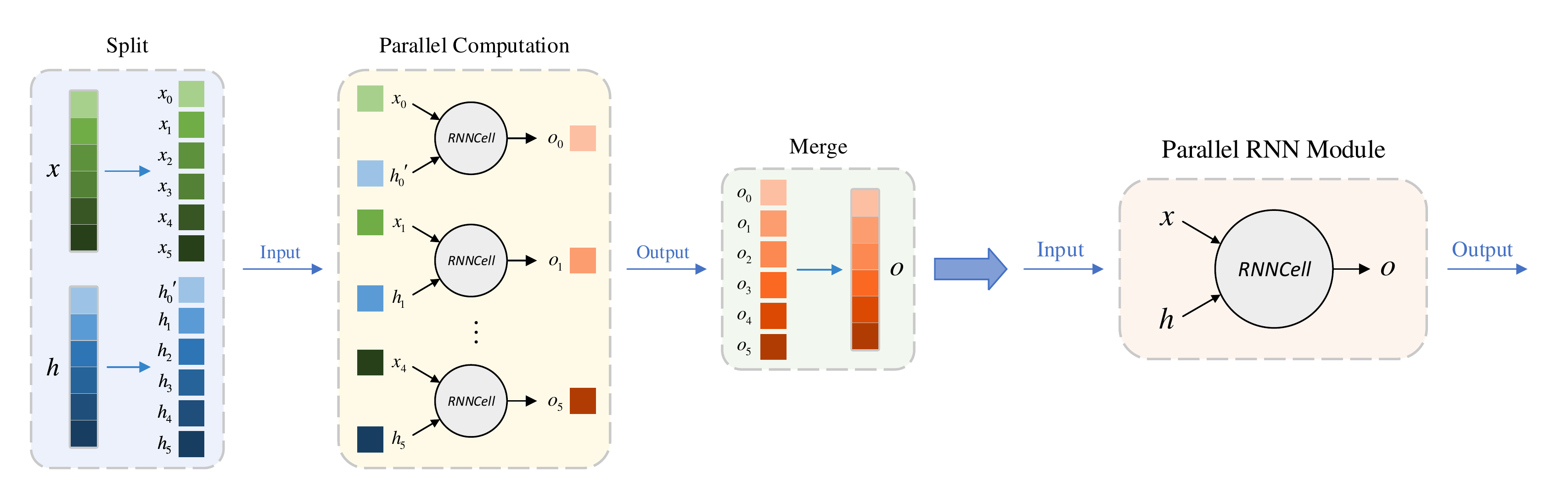}
	\caption{\textbf{Part Left} shows the operations of parallel RNN computation. By carry-lookahead module, hidden states $h$ are calculated in advance. The input sequence $x$ and hidden states $h$ are split to word vectors. Word vector pair $(x_i, h_i)$ is fed to a \emph{RNNCell} as input and all \emph{RNNCell}s run in parallel. The outputs are merged in the end. \textbf{Part Right} shows the structure of parallel RNN module. \emph{RNNCell}s are integrated to one \emph{RNNCell}. The whole length of sequence is calculated as an entirety.}\label{RNN-integration}
\end{figure*}

The classic RNN (Fig.~\ref{RNN}) can be viewed as consisting of a series of parameter-shared \emph{RNNCell}s. Each \emph{RNNCell} receives the hidden state $h_i$ with the corresponding word vector $x_i$ to produce the output $o_i$. The formula is shown as:
\begin{equation}\label{RNNcell}
	o_i = tanh(x_iW_{ih} + b_{ih} + h_iW_{hh} + b_{hh})
\end{equation}
where $W_{ih}$, $W_{hh}$, $b_{ih}$ and $b_{hh}$ are internal trainable parameters.

Limited by serial dependency, RNNs can only process on a word-by-word mode, because each \emph{RNNCell} must wait for the completion of the previous one to obtain the hidden state, then to generate the next one.

In the last section, through carry-lookahead module, hidden states $h$ are calculated in advance, which makes it possible for RNN to run in parallel. In this section, we transform the classic RNN into a parallel version and give the detailed method.

Parallel RNN computation is implemented vis three operations - \emph{Split}, \emph{Parallel computation} and \emph{Merge}, as illustrated in Fig.~\ref{RNN-integration} (Part Left). RNN deals with sequence in word-level, so the input sequence $x$ is split to word vectors $x_0, x_1, \cdots, x_n$, and $h$ is also split correspondingly. Each \emph{RNNCell} takes the respective $(x_i, h_i)$ as input and runs in parallel. The outputs are merged in the end.

Taking the input sequence "I love the bird" as an example.

\textbf{\emph{Input preprocess:  }}Each word in the sequence is first encoded to numeric vector (word vector) by embedding layer~\cite{le2014distributed}. In embedding layer, the word is converted to an unique value based on dictionary order, then encoded to one-hot vector and mapped to word vector by linear projection. The internal parameters in embedding layer can be optimized with stochastic gradient descent (SGD) to better represent the words. When embedding size is set to 100, each word vector has 100 elements.
\[x = Embedding\left( \begin{array}{l}
	I\\
	love\\
	the\\
	bird
	\end{array} \right) = \left( \begin{array}{l}
	{\rm{0}}{\rm{.014, 0}}{\rm{.000, }} \cdots {\rm{,  - 0}}{\rm{.011}}\\
	{\rm{ - 0}}{\rm{.008,  - 0}}{\rm{.000, }} \cdots {\rm{, 0}}{\rm{.000}}\\
	{\rm{ - 0}}{\rm{.006,  - 0}}{\rm{.002, }} \cdots {\rm{, 0}}{\rm{.006}}\\
	{\rm{0}}{\rm{.002, 0}}{\rm{.018, }} \cdots {\rm{,  - 0}}{\rm{.003}}
\end{array} \right)\]
Here $x$ is formed by the four word vectors and the shape is $(4, 100)$.

Then, $x$ is sent to carry-lookahead module (CL\underline{~~}Module) to generate the hidden states in advance.

\[h = CL\underline{~~}Module(h_0, x) = \left( \begin{array}{l}
	{\rm{ - 0}}{\rm{.010, 0}}{\rm{.009, }} \cdots {\rm{, 0}}{\rm{.067}}\\
	{\rm{ - 0}}{\rm{.024, 0}}{\rm{.024, }} \cdots {\rm{, 0}}{\rm{.064}}\\
	{\rm{ - 0}}{\rm{.018, 0}}{\rm{.009, }} \cdots {\rm{, 0}}{\rm{.084}}\\
	{\rm{ - 0}}{\rm{.017, 0}}{\rm{.028, }} \cdots {\rm{, 0}}{\rm{.071}}
\end{array} \right)\]

\textbf{\emph{Split:  }}The input sequecne $x$ and hidden states $h$ are split to separate word vectors.

\[\left\{ \begin{array}{l}
	x = ({x_0},{x_1},{x_2},{x_3})\\
	h = ({h_0}^\prime,{h_1},{h_2},{h_3})
\end{array} \right.\]

\textbf{\emph{Parallel computation:  }}Corresponding word vectors form the input pair $(x_i, h_i)$ and sent to a \emph{RNNCell}. There is no dependency between \emph{RNNCell}s, so they can calculate in parallel.

\[\left\{ \begin{array}{l}
	{o_0} = RNNCell({x_0},{h_0}^\prime ) = ({\rm{ - 0}}.{\rm{210}},{\rm{0}}.{\rm{035}}, \cdots ,{\rm{0}}.{\rm{036)}}\\
	{o_1} = RNNCell({x_1},{h_1}) = ({\rm{ - 0}}.{\rm{064}},{\rm{0}}.{\rm{034}}, \cdots ,{\rm{0}}.{\rm{044}})\\
	{o_2} = RNNCell({x_2},{h_2}) = ({\rm{0}}.{\rm{035}},{\rm{ - 0}}.{\rm{048}}, \cdots ,{\rm{0}}.{\rm{078}})\\
	{o_3} = RNNCell({x_3},{h_3}) = ({\rm{0}}.{\rm{036}},{\rm{0}}.{\rm{067}}, \cdots ,{\rm{ - 0}}.{\rm{041}})
\end{array} \right.\]

\textbf{\emph{Merge:  }}Finally, the outputs of \emph{RNNCell}s $\{{o_0},{o_1},{o_2},{o_3}\}$, which are vectors with 100 elements respectively, are merged in the end. The shape of $o$ is $(4, 100)$, which has the same length as $x$.

\[({o_0},{o_1},{o_2},{o_3}) = o\]

\emph{RNNCell} in the classic RNN, limited by its serial dependency, has to deal with the input $x$ word by word. While, with the proposed carry-lookahead module, hidden states $h$ are calculated in advance, so the input $x$ can be used to calculate as an entirety. Thus, \emph{RNNCell} in parallel RNN module, shown as Fig.~\ref{RNN-integration} (Part Right), considers the sequence $x$ and $h$ as entireties. The formula is expressed as:
\begin{equation}\label{RNNCell}
	o = RNNCell(x, h) = tanh(xW_{ih} + b_{ih} + hW_{hh} + b_{hh})
\end{equation}
where $W_{ih}$, $W_{hh}$, $b_{ih}$ and $b_{hh}$ refer to the same things as in Equation (\ref{RNNcell}).

In Equation (\ref{RNNcell}), which shows the principle of RNNCell in the classic RNN, $x$ is separately dealt as $\{{x_0},{x_1}, \cdots ,{x_n}\}$. In Equation (\ref{RNNCell}), $x$ and $h$ are considered as entireties, which is further integrated.

Meanwhile, from a global perspective, parallel RNN module plays a role as residual operation to carry-lookahead module (show as the red line in Fig.~\ref{Model-structure}). Residual operation is introduced to neural network by He et al.~\cite{he2016deep} to ease the model optimization difficulty and it now has been widely applied to image classification, image dehazing, object detection, etc.~\cite{wang2017residual, zhong2017spectral, zhang2020drcdn, chen2020reverse, zhang2020weight} Compared with the classic residual operation, using parallel RNN module as residual component has following characteristics: 

\begin{enumerate}
	\item[1)] Compound residual operation: parallel RNN module is a generalization of the classic residual operation, which adds linear projection and bias to inputs. When all internal parameters are initialized to identity matrices, it degenerates to the classic residual operation.
	\item[2)] Trainability of residual operation: the parameters in parallel RNN module are optimized through iteration. The optimized residual operation makes the model have stronger representation capability.
	\item[3)] Merging of any dimension: the input sequence will have the same shape as the hidden state after the linear projection and can be added directly. There is no need to design a special padding layer or convolution layer for specific dataset to increase or reduce the dimension, which improves its generalizability.
\end{enumerate}

Due to the introduction of linear projection, parallel RNN module takes more parameters than the classic residual operation. However, since CL-RNN is not large in model size, it is acceptable to appropriately increase the number of parameters to improve the model performance, which is verified by ablation experiment in Sec.~\ref{sec-experiment-sMNIST}.

\subsection{Evaluation of the Proposed Model}

Using CL-RNN for sequence modeling has the following characteristics.

\begin{itemize}
	\item \textbf{Parallelism.} CL-RNN break the serial dependency of RNN by introducing carry-lookahead module to calculate hidden states in advance, which makes it possible for RNN to run in parallel. Based on that, parallel RNN module specifically achieves parallel RNN computation.
    \item \textbf{Flexible receptive field.} The receptive field of neurons in carry-lookahead module is exponential expanded with the number of layers and can be further enhanced by dilated convolution. As a result, a very large receptive field can be achieved with a few layers.
	\item \textbf{Compound residual operation.} Parallel RNN module plays a role as residual operation in the global model. It improves the representation
	capability and eases the model optimization difficulty, which avoids the model being restricted to local optimal solutions.
\end{itemize}

\section{Experiment}\label{sec-experiment}

In this section, we validate the performance of CL-RNN in sequence modeling on 5 representative datasets. The models are implemented with PyTorch and trained in Google Colab with NVIDIA Tesla T4 GPU.

The MNIST database is one of the most common databases used for model testing. Sequential MNIST is one of its varietas, which is specially designed to test sequence models. To validate the effectiveness of modules in CL-RNN, ablation experiment is executed on Sequential MNIST first. Then, the model is tested in character-level language modeling on PTB and text8 datasets, which is a classic problem in natural language processing. Finally, to test the generalization ability of the model, polyphonic music modeling is executed on JSB Chorales and Nottingham datasets, where CL-RNN shows leading advantage.

If not specified, all models are trained through 100 epochs with learning rate decay~\cite{NIPS1990_75887499}. Learning rate decay is a dynamic learning rate strategy, which can help model to converge better. When the test loss is greater than the maximum of the last three times, the learning rate will be divided by 10.

\subsection{Ablation Experiment on Sequential MNIST}\label{sec-experiment-sMNIST}

\begin{table}
	\centering
	\caption{The hyperparameters on Sequential MNIST.}
	\renewcommand\arraystretch{1.2}
	\begin{tabular}{l|r}
		\hline
		Hyperparameter        & Value       \\ \hline
		Batch size            & 64          \\
		Epochs                & 12          \\
		Initial learning rate & 2e-3        \\
		Kernel size           & 7           \\
		Number of layers      & 8           \\
		Number of kernels (per layer) & 1   \\
		Optimizer             & Adam        \\
		Dropout               & 0.05        \\ \hline
	\end{tabular}
	\label{tab-test-hyper-sMNIST}
\end{table}

\begin{table}
	\centering
	\caption{Evaluation of models on Sequential MNIST on test set.}
	\renewcommand\arraystretch{1.2}
	\begin{tabular}{l|c|c}\toprule[2pt]
		Model & Top-1 Accuracy & Model Size ($\approx$)\\ \hline
		CL-RNN & \textbf{98.02\%} & 1M\\
		TCN~\cite{lea2017temporal} & 11.35\% & 8K\\
		TCN variant~\cite{bai2018empirical} & 94.43\% & 10K\\
		RNN~\cite{rumelhart1986learning} & 92.83\% & 8K\\
		LSTM~\cite{hochreiter1997long} & 93.68\% & 8K\\
		\midrule[2pt]
	\end{tabular}
	\label{tab-test-sMNIST}
\end{table}

The MNIST database~\cite{lecun1998gradient} is a large database of handwritten digits, which is the most common database used in image classification. Sequential MNIST is based on the MNIST, but each image (size: $28 \times 28$) is represented as a $784 \times 1$ sequence. It is adopted by a great deal of researches to test the ability of recurrent neural networks to retain past information~\cite{le2015simple, zhang2016architectural, wisdom2016full, kalchbrenner2014convolutional, krueger2016zoneout, jing2017tunable}.

All the neural network models are not pre-trained, and the hyperparameters are set as shown in Table~\ref{tab-test-hyper-sMNIST}, where epochs are set to 12, due to the overfit of models when epochs exceed 12.

It is worth noting that in order to show the differences between models, control variable method, instead of grid search method, is adopted for hyperparameters. Otherwise, if the model size is violently enlarged, such as setting the number of convolution kernels in the hidden layer to 25, all CL-RNNs are able to reach 99\% accuracy, which makes models lose their comparative meaning. So, we adopt control variable method to highlight the impact of the adjustment to the models.

The results of ablation experiment on Sequential MNIST are shown in Table~\ref{tab-test-sMNIST-ae} and the architectures of the involved models are shown in Table~\ref{tab-architecture-sMNIST}. Through the comparison between CL-RNN(1)\underline{~~}shortcut and CL-RNN(1), it can be seen that the performance of CL-RNN is greatly improved after applying parallel RNN module as residual component while the model size is similar, which proves the effectiveness of parallel RNN module.

\begin{table*}[htbp]
	\centering
	\caption{Ablations of modules on Sequential MNIST. CL-RNN(1) refers to the linear layer (at the end of the model) uses only the last item of parallel RNN module's output sequence; while CL-RNN(784) uses the whole output sequence instead; CL-RNN(1)\underline{~~}shortcut refers to the network that parallel RNN module is replaced by addition operation, which is adopted in classic residual operation; addCNN/addRNN refers to CL-RNN with 1D-CNN or RNN layer added behide parallel RNN module.}
	\renewcommand\arraystretch{1.2}
	\begin{tabular}{lcc}\toprule[2pt]
		Model & Top-1 Accuracy & Model Size ($\approx$)\\
		\midrule
		CL-RNN(1)\underline{~~}shortcut & 92.58\% & 8K\\
		\textbf{CL-RNN(1)} & 94.72\% & 8K\\
		CL-RNN(784) & \textbf{98.02\%} & 1M\\
		CL-RNN(784)\underline{~~}ReLU & 97.71\%) & 1M\\
		CL-RNN(784)\underline{~~}Dropout(0.2) & \textbf{98.05\%} & 1M\\
		CL-RNN(784)\underline{~~}ReLU+Dropout & 97.72\% & 1M\\
		CL-RNN(784)\underline{~~}ReLU+Dropout\underline{~~}addCNN & 96.51\% & 1M\\
		CL-RNN(784)\underline{~~}addRNN & 97.87\% & 1M\\
		\midrule[2pt]
	\end{tabular}
	\label{tab-test-sMNIST-ae}
\end{table*}

\begin{table*}[htbp]
	\centering
	\caption{Architectures of the models for Sequential MNIST. Inside the brackets are the general shape of dilated causal convolutional blocks, including filter sizes and feature dimensionalities. $d$ denotes dilation factor, $p$ denotes padding, $s$ denotes stride, and $i$, $h$ are the feature dimensionalities of input and output respectively.}
	\scalebox{0.80}{
	\renewcommand\arraystretch{1.2}
	\begin{tabular}{c|cccc}
	\hline
	Output size & \multicolumn{1}{c|}{CL-RNN(1)} & \multicolumn{1}{c|}{CL-RNN(784)} & \multicolumn{1}{c|}{CL-RNN(784)\underline{~~}addCNN} & CL-RNN(784)\underline{~~}addRNN\\ \hline
	$784 \times 1$ & \multicolumn{4}{c}{$\left[ \begin{array}{l}
		7 \times 7,1,d = 1,p = 6\\
		7 \times 7,1,d = 2,p = 12\\
		7 \times 7,1,d = 4,p = 24\\
		7 \times 7,1,d = 8,p = 48\\
		7 \times 7,1,d = 16,p = 96\\
		7 \times 7,1,d = 32,p = 192\\
		7 \times 7,1,d = 64,p = 384\\
		7 \times 7,1,d = 128,p = 768
	\end{array} \right]$}\\ \hline
	$784 \times 1$ & \multicolumn{1}{c|}{$RNNCell(i=1, h=1)$} & \multicolumn{3}{c}{$RNNCell(i=784, h=784)$}\\ \hline
	$784(23) \times 1$ & \multicolumn{2}{c|}{Identity} & \multicolumn{1}{c|}{$\left[ \begin{array}{l}
		5 \times 5, 1, s = 5, p = 1\\
		7 \times 7, 1, s = 7, p = 2
	\end{array} \right]$} & $RNN(i=1, h=1)$\\ \hline
	$10 \times 1$ & \multicolumn{2}{c|}{784-d $fc$, log\underline{~~}softmax} & \multicolumn{1}{c|}{23-d $fc$, log\underline{~~}softmax} & 784-d $fc$, log\underline{~~}softmax\\ \hline
	\end{tabular}}
	\label{tab-architecture-sMNIST}
\end{table*}

\begin{figure*}
	\centering
	\includegraphics[width = 0.65\textwidth]{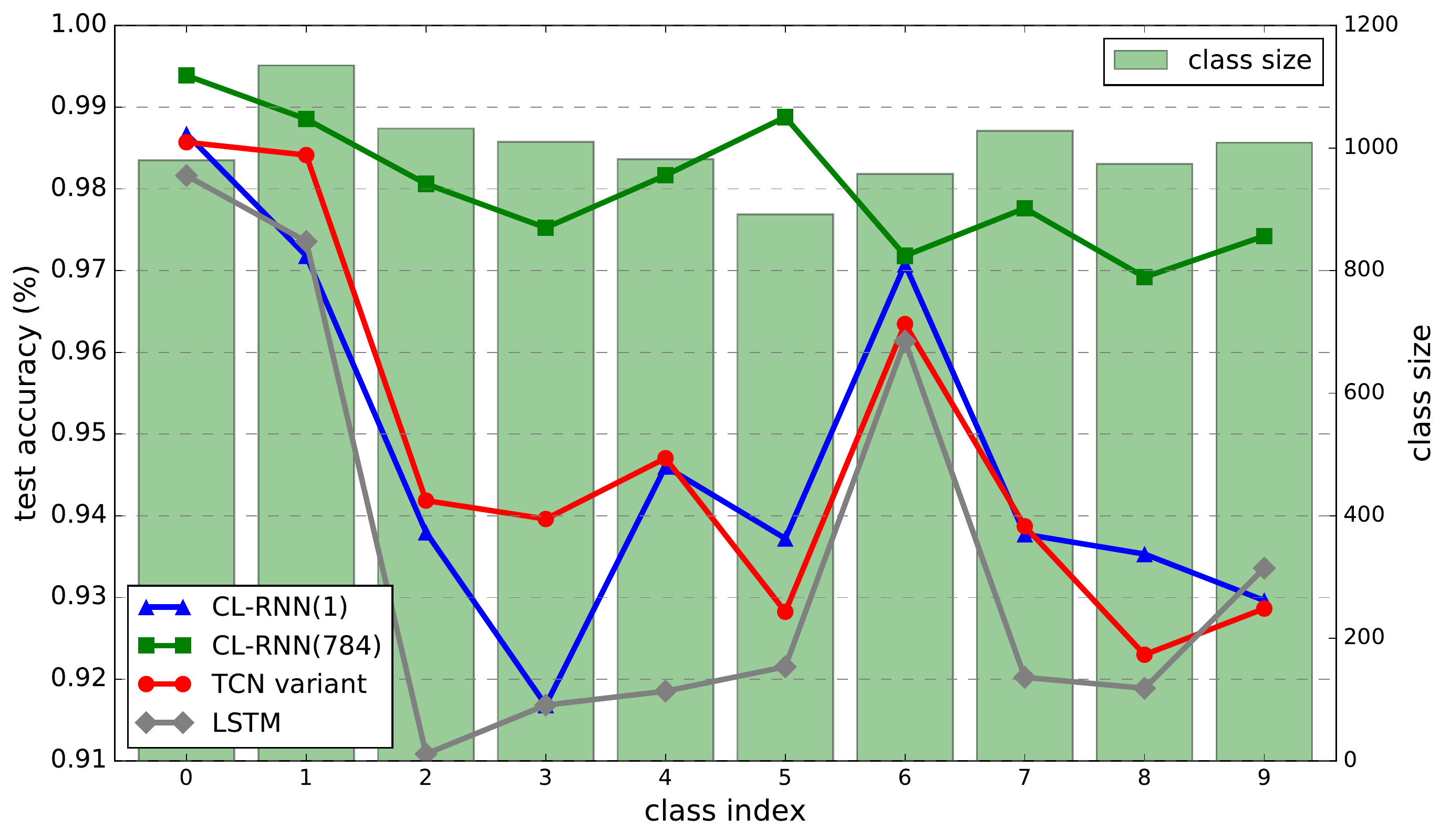}
	\caption{Comparison of model test accuracy on Sequential MNIST in the 10 classes. Class size denotes the number of items in this class on test set. Total size of the 10 classes is 10,000.}\label{class_accuracy}
\end{figure*}

\begin{figure*}[htbp]
	\centering
	\subfigure[\label{char_ptb}]{\includegraphics[width = 0.46\textwidth]{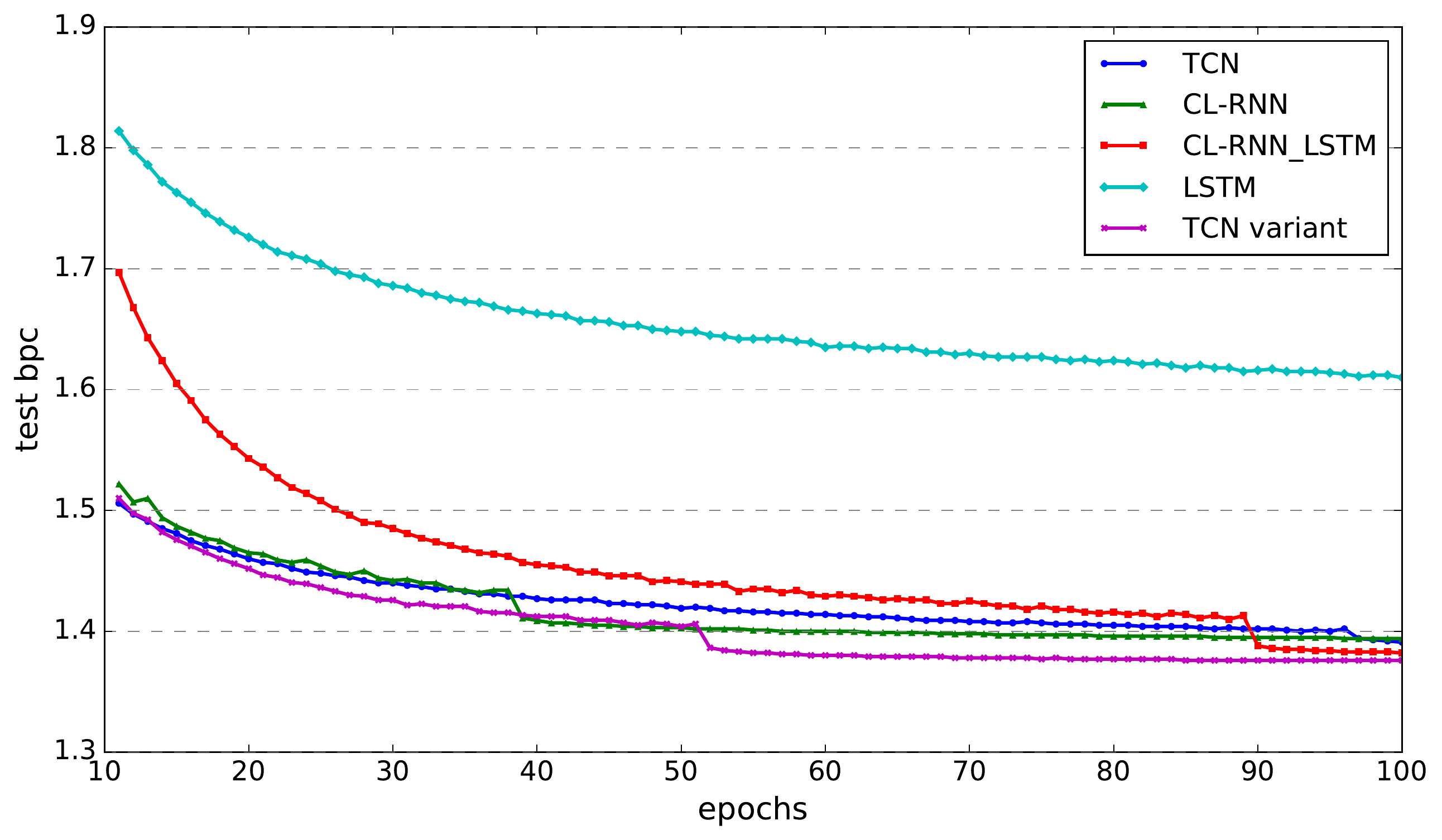}}
	\subfigure[\label{char_text8}]{\includegraphics[width = 0.47\textwidth]{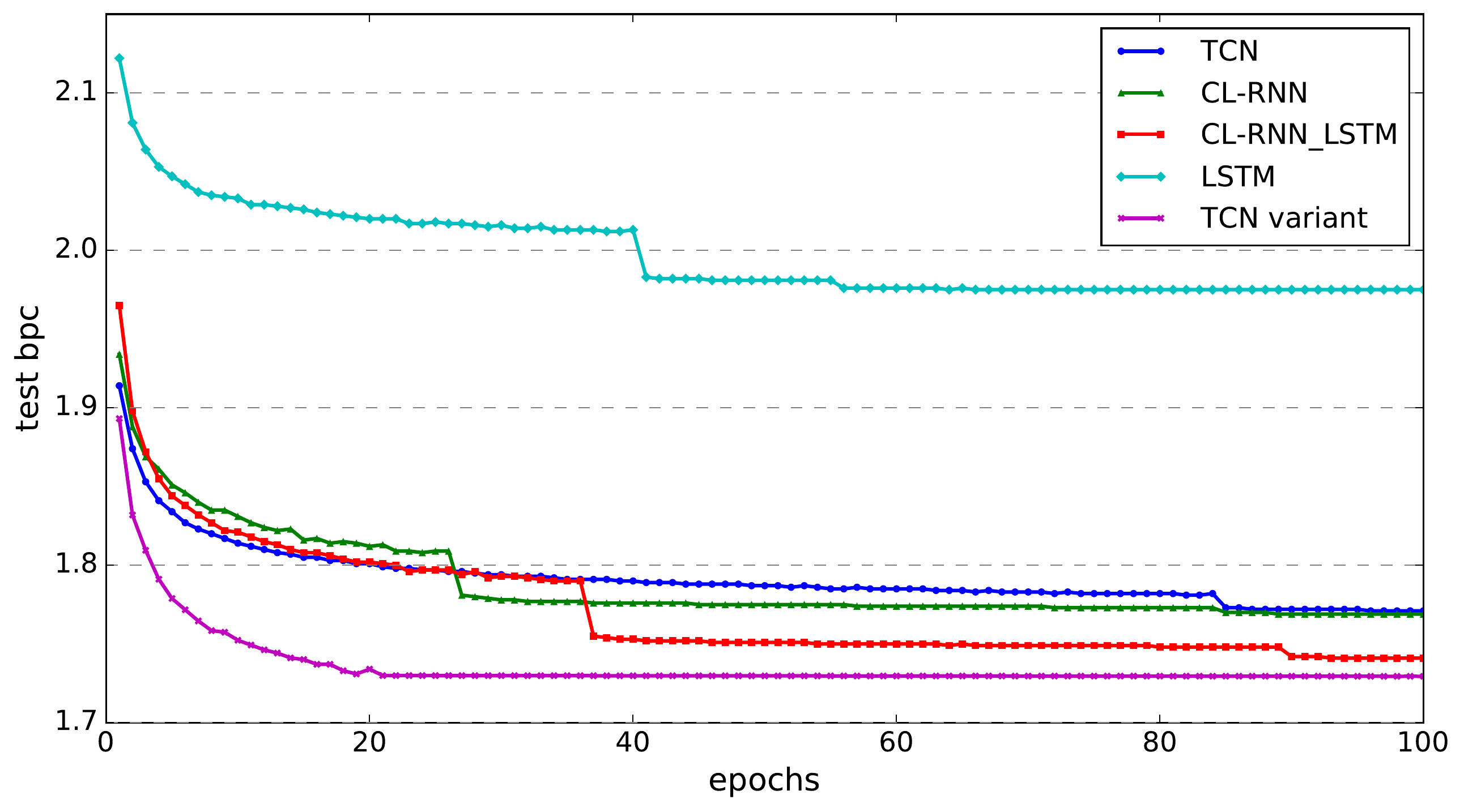}}
	\caption{Comparison of model bpc on (a) PTB and (b) text8 on test set.}
\end{figure*}

The last item of output sequence is generated by the neuron with the widest receptive field. So CL-RNN(1) uses only the last item can achieve such good results. While CL-RNN(784) uses the entire output sequence for classification, which means the result is calculated based on multiple scales of input information. Although it brings much more parameters to linear layer making the model size bigger, the accuracy further increases to 98\%, which is significantly improved. To balance the performance and efficiency, CL-RNN(1) is the comprehensive optimal choice; to aim at the higher accuracy, CL-RNN(784) is the best choice.

Fig.~\ref{class_accuracy} shows the distribution of the test set and the test accuracy of the models in the 10 classes. All models perform well in the classes of digit with simple stroke, such as 0 and 1. But when the stroke are relatively complex, such as 2 and 8, the accuracy fluctuates considerably, which can well reflect the representation capability of the models. CRN(784) not only perform best in each class, but also has the smoothest curve, which indicates its representation capability is sufficient for the complex digits.

Some other modules are also tried to be added on the model. Since the output sequence of parallel RNN module is as same length as the input sequence. It's convenient to add 1D-CNN or RNN layer behind it. The results show no significant improvement in accuracy, which indicates that CL-RNN(1) and CL-RNN(784) are suitable enough on Sequential MNIST. Therefore, in the subsequent experiments, CL-RNN(784) is used as the main model.

In comparison experiment, CL-RNN is compared with TCN, RNN, LSTM. The results are shown in Table~\ref{tab-test-sMNIST}. Considering the models are trained from scratch on the same dataset, the results can objectively verify the performance and feasibility of CL-RNN. According to Table~\ref{tab-test-sMNIST}, CL-RNN performs much better than the classic sequence models.

Besides, it's found that TCN performs quite bad in the tasks. We suppose the reason is that the network can't be well initialized in this task, which makes the model restricted to a local optimal solution and can't get further optimized. Although several methods (e.g. change the initialization methods, change the random seed, change the initial learning rate) have been tried, the results remain similar. Compared with that, Parallel RNN Module in our proposed CL-RNN plays a role as the residual component in the network, which eases the optimization difficulty and avoids such problem. As a result, the performance gets significant improvement.

As TCN is unable to normally perform, CL-RNN is also compared with the variant of TCN. According to the results, CL-RNN has better performance, which further verifies the effectiveness of the proposed model.

\subsection{Character-level Language Modeling}

\begin{table}
	\centering
	\caption{The hyperparameters on character-level language modeling.}
	\renewcommand\arraystretch{1.2}
	\begin{tabular}{l|r}
		\hline
		Hyperparameter        & Value         \\ \hline
		Batch size            & 32            \\
		Epochs                & 100           \\
		Kernel size           & 3             \\
		Number of layers      & 3             \\
		Number of kernels (per layer) & 450   \\
		Optimizer             & SGD           \\
		Dropout               & 0.1           \\
		Gradient Clip         & 0.15          \\
		Initial Learning rate & 4             \\
		Embedding size        & 100           \\
		Valid sequence length & 320           \\
		Sequence length       & 400           \\ \hline
	\end{tabular}
	\label{tab-test-hyper-char}
\end{table}

\begin{table}
	\centering
	\caption{Evaluation of models on character-level language modeling on test set. CL-RNN\underline{~~}LSTM refers to CL-RNN with LSTM added behind parallel RNN module.}
	\renewcommand\arraystretch{1.2}
	\setlength{\tabcolsep}{3.5pt}
	\begin{tabular}{l|c|c|c}\toprule[2pt]
		Model & bpc (PTB) & bpc (text8) & Model Size ($\approx$)\\
		\hline
		CL-RNN       & 1.394     & 1.769       & 0.9M   \\
		CL-RNN\underline{~~}LSTM & 1.382     & 1.741       & 1M     \\
		TCN~\cite{lea2017temporal}       & 1.391     & 1.771       & 0.9M   \\
		TCN variant~\cite{bai2018empirical} & \textbf{1.376} & \textbf{1.730} & 3M   \\
		LSTM~\cite{hochreiter1997long}      & 1.587     & 1.975       & 85K    \\
		\midrule[2pt]
	\end{tabular}
	\label{tab-test-char}
\end{table}

\begin{figure*}[htbp]
	\centering
	\subfigure[\label{music_JSB}]{\includegraphics[width = 0.48\textwidth]{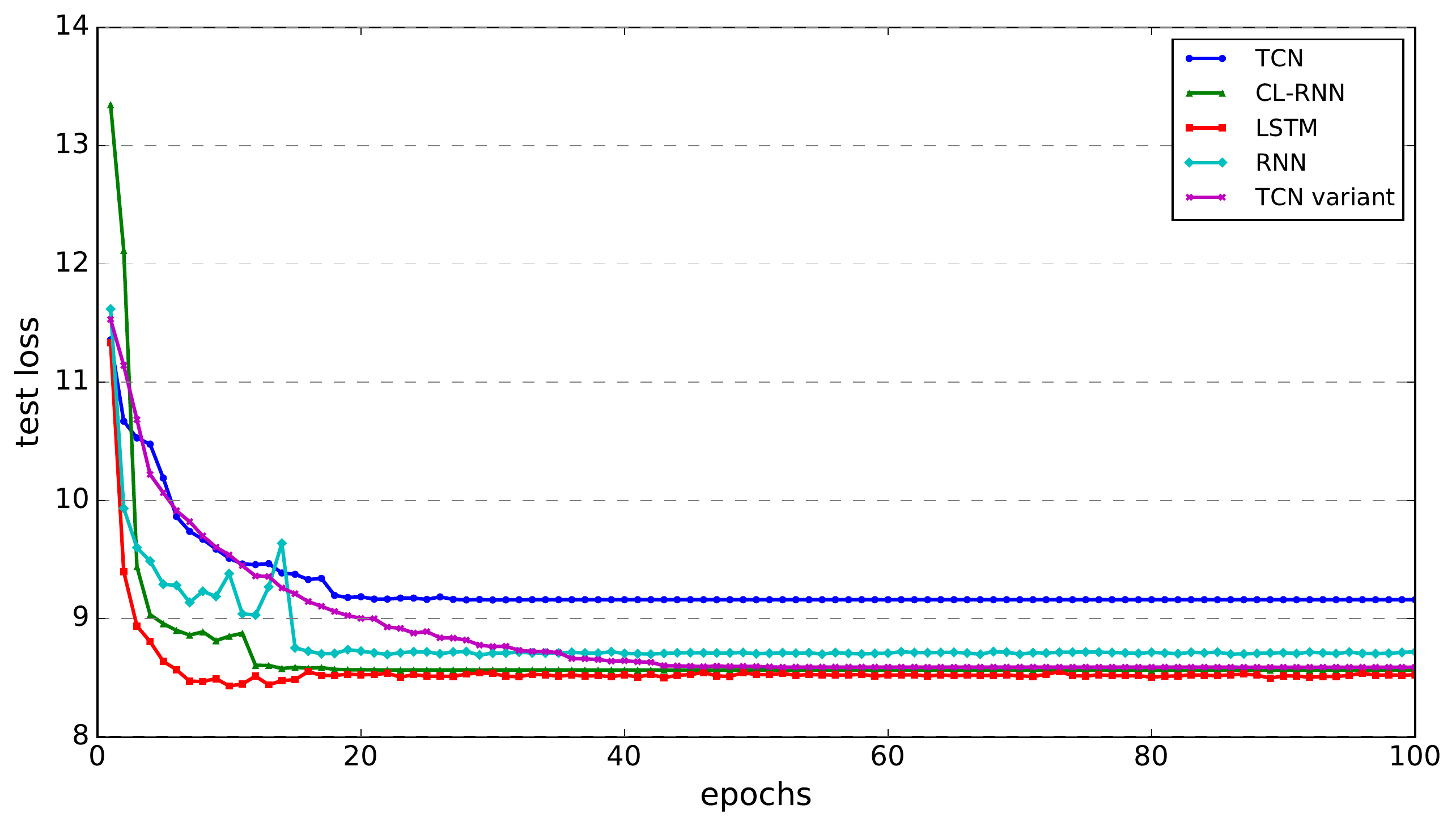}}
	\subfigure[\label{music_Nott}]{\includegraphics[width = 0.48\textwidth]{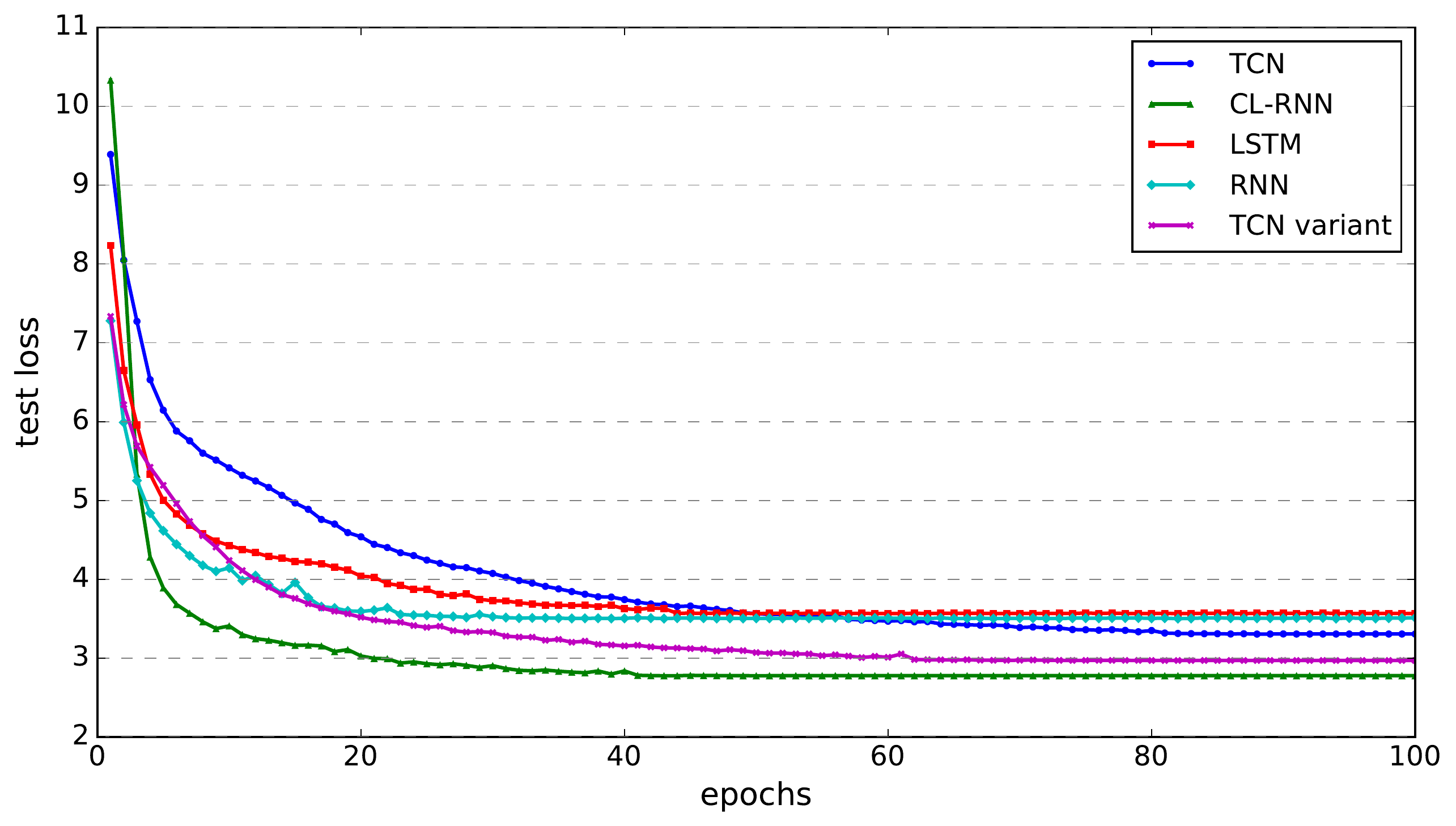}}
	\caption{Comparison of model loss on (a) JSB and (b) Nott on test set.}
\end{figure*}

Character-level language modeling treats each character in text as a word, which greatly compresses the encoding space. Compared with word-level language modeling, it ignores the semantics of words in the text but requires the model to capture the semantic information. The idea of character-level language modeling comes from signal processing and it’s a classic problem in natural language processing. This paper makes a comparison test of character-level language modeling on two datasets.

\begin{itemize}
	\item PennTreebank (PTB)\\ PennTreebank~\cite{marcus1993building} consists of 49 characters and contains 5059K characters for training, 396K characters for validation, and 446K characters for test. It’s a relatively small but highly studied dataset~\cite{miyamoto2016gated, krueger2016zoneout, merity2017regularizing}.
	\item text8\\ text8~\cite{mikolov2012subword} consists of 27 characters and has about 20 times the amount of data as PennTreebank. It contains 100M characters from Wikipedia, 90M of which are used for training, 5M for validation, and 5M for test.
\end{itemize}

In implementation, \emph{Sequence length} characters are intercepted from the dataset as a sentence. For forward propagation, the whole sentence is used as the input sequence. For backward propagation, only the last \emph{Valid sequence length} characters of the sentence are used, which ensures that the training data are also provided with sufficient history.

We use cross-entropy as the loss function and bits-per-character (bpc)~\cite{graves2013generating} to measure the performance of the language model, which is an usual measure index in language modelling. Bpc is calculated with the following formula, and it can be seen that the lower bpc is better.

\begin{equation}
	bpc = \frac{{\overline {loss} }}{{\ln (2)}}
\end{equation}

The hyperparameters of the two tests are set as shown in Table~\ref{tab-test-hyper-char}, and the test results are shown in Table~\ref{tab-test-char}. Fig.~\ref{char_ptb} and Fig.~\ref{char_text8} show the loss of model versus iteration on the two datasets.

According to Table~\ref{tab-test-char}, it's found that the performance of LSTM is much worse than other models, probably due to the small number of parameters which limits the representation capability of the model. Besides, the performance of TCN and CL-RNN is similar.

Although the performance of CL-RNN and LSTM itself is mediocre, CL-RNN\underline{~~}LSTM, which is formed by the superposition of the two, is much better. It achieves the lowest bpc in both two tests. Moreover, Fig.~\ref{char_ptb} and Fig.~\ref{char_text8} shows that CL-RNN\underline{~~}LSTM is not fully converged after 100 epochs of training, which indicates that there is still room for further optimization.

With three times the size than other models, TCN variant achieves better performance in character-level language modeling. However, big model takes long time to train. Compared with that, CL-RNN\underline{~~}LSTM well balances the relationship between the model size and performance.

\subsection{Polyphonic Music Modeling}

\begin{table}
	\centering
	\caption{The hyperparameters on polyphonic music modeling.}
	\renewcommand\arraystretch{1.2}
	\begin{tabular}{l|r}
		\hline
		Hyperparameter               & Value         \\ \hline
		Epochs                       & 100           \\
		Kernel size                  & 5             \\
		Number of layers             & 4             \\
		Number of kernels (per layer) & 150          \\
		Optimizer                    & Adam          \\
		Dropout                      & 0.25          \\
		Gradient Clip                & 0.2           \\
		Initial Learning rate (JSB)  & 1e-2          \\
		Initial Learning rate (Nott) & 1e-3          \\ \hline
	\end{tabular}
	\label{tab-test-hyper-poly-music}
\end{table}

\begin{table}
	\centering
	\caption{Evaluation of models on polyphonic music modeling on test set.}
	\renewcommand\arraystretch{1.2}
	\scalebox{0.9}{\begin{tabular}{l|c|c|c}\toprule[2pt]
		Model & Loss (JSB) & Loss (Nott) & Model Size ($\approx$)\\
		\hline
		CL-RNN   & 8.565    & \textbf{2.776}     & 455K   \\
		TCN~\cite{lea2017temporal}   & 9.158    & 3.305     & 418K   \\
		TCN variant~\cite{bai2018empirical} & 8.589 & 2.969 & 883K   \\
		RNN~\cite{rumelhart1986learning}   & 8.707   & 3.506   & 50K   \\
		LSTM~\cite{hochreiter1997long}   & \textbf{8.442}   & 3.567   & 157K   \\
		\midrule[2pt]
	\end{tabular}}
	\label{tab-test-poly-music}
\end{table}

A piece of music consists of a sequence of notes, and many researches have applied recurrent network architecture to music modeling tasks~\cite{chung2014empirical, pascanu2013construct, jozefowicz2015empirical, greff2016lstm}. In this paper, CL-RNN is tested on two polyphonic music datasets.

\begin{itemize}
	\item JSB Chorales (JSB)\\ JSB Chorale~\cite{allan2005harmonising} is a polyphonic music dataset consisting of 382 four-part harmonic works composed by J. S. Bach. Each element in the input sequence consists of 88 bits, corresponding to the 88 keys on the piano, and value of 1 indicates that the key is pressed at that moment.
	\item Nottingham (Nott)\\ Nottingham is a dataset consisting of 1200 English and American folk tunes, which is a larger dataset compared to JSB Chorales.
\end{itemize}

The hyperparameters of the two tests are set as shown in Table~\ref{tab-test-hyper-poly-music}, except that the initial learning rates for TCN on Nott dataset and Residual TCN on JSB and Nott datasets are set to 1e-4 (gradient explosion occurs at 1e-3). Cross-entropy is used as the loss function and the value of loss on test set is used to measure the performance of the models. We compare CL-RNN with TCN, RNN, and LSTM, and the results are shown in Table~\ref{tab-test-poly-music}. Fig.~\ref{music_JSB} and Fig.~\ref{music_Nott} show the loss of models versus iteration.

According to Table~\ref{tab-test-poly-music}, CL-RNN performs well in both tests. Although it is just slightly inferior to LSTM on JSB dataset, it is far ahead compared to other models on Nott dataset, which shows the strong representation and generalization capability.

Furthermore, as shown in Fig.~\ref{music_JSB} and especially Fig.~\ref{music_Nott}, CL-RNN converges significantly faster than the other models and achieves good result at the early stage of training, which indicates the model is well and easily optimized on this dataset.

\section{Conclusion and Future Work}\label{sec-conclusion}

Considering the analogy between serial adder and RNN, this paper puts forward the idea of applying mature optimization experience on adders to improving RNN. Carry-lookahead adder is the mainstream adder at present. Based on its core idea, CL-RNN is proposed, which consists of carry-lookahead module and parallel RNN module. It overcomes the problem of serial dependency of recurrent architecture and has the advantages of parallelism, flexible receptive field, compound residual operation, etc. The proposed CL-RNN specifically targets a comprehensive set of sequence modeling tasks that have been repeatedly used to compare the effectiveness of different recurrent neural networks. Experimental results prove that CL-RNN can perform better than classic TCN, RNN and LSTM in this domain.

The core idea of this paper is the transfer of adder optimization experience and the proposal of CL-RNN, but deeper stack of CL-RNN is not explored. Referring to the architecture of ResNet~\cite{he2016deep}, the core part of CL-RNN can be viewed as an integrated RNN residual block to build deeper networks in the future. Besides, more variants of TCN have been proposed as TCN gradually gets popular in sequence modeling tasks. Meanwhile, Carry-lookahead RNN based on various variants of TCN can be further explored.

\section*{Acknowledgement}

This work was supported by National Natural Science Foundation of China (Nos. 61972121, 61971173, U1909210), Zhejiang Provincial Natural Science Foundation of China (Nos. LY21F020015, LY20F020015, LY21F030005), Fundamental Research Funds for the Provincial Universities of Zhejiang (No. GK209907299001-008), National Social Science Foundation of China (No. 19ZDA348). The authors would like to thank the reviewers for their comments and suggestions in advance.

%% \section*{References}\label{sec-ref}


%% References



\bibliographystyle{elsarticle-num}
\bibliography{cellDL}

%% not want to use model3-num-names.bst.


%% \begin{thebibliography}{00}

%% \bibitem{key}...


% \end{thebibliography}

\end{sloppypar}
\end{document}